\documentclass[10pt,twocolumn,letterpaper]{article}

\usepackage{cvpr}              
\definecolor{cvprblue}{rgb}{0.21,0.49,0.74}
\usepackage[pagebackref,breaklinks,colorlinks,allcolors=cvprblue]{hyperref}
\usepackage{microtype}
\usepackage{graphicx}
\usepackage{booktabs} 
\usepackage{multirow}
\usepackage{lipsum}  
\usepackage{caption} 
\usepackage{subcaption}
\usepackage{colortbl}
\DeclareUnicodeCharacter{2061}{}
\usepackage{xspace}
\usepackage{xcolor}

\def\HyQuant{EfficientQuant\xspace}

\def\logtbas{$\log_{2}$\xspace}
\def\search{structure-aware\xspace}
\usepackage[ruled,vlined]{algorithm2e}

\usepackage{amsmath}
\usepackage{amssymb}
\definecolor{keywordcolor}{rgb}{0.0, 0.0, 0.5}

\SetKwComment{Comment}{/* }{ */}
\SetCommentSty{itshape}
\definecolor{cblue}{RGB}{16,78,139} 
\definecolor{cred}{RGB}{139,37,0}   

\newcommand{\highlight}[1]{\textcolor{black}{#1}}
\def\paperID{10} 
\def\confName{CVPR}
\def\confYear{2025}
\AddToShipoutPicture*{%
    \AtPageUpperLeft{%
        \put(64,-60){\parbox{\textwidth}{
        \textcolor{blue}{\underline{Accepted to the 4th Workshop on Transformers for Vision (T4V) at CVPR 2025}}\\[0.5ex]
        }}
    }
}
\title{EfficientQuant: An Efficient Post-Training Quantization for CNN-Transformer Hybrid Models on Edge Devices}
\author{Shaibal Saha \quad Lanyu Xu\\
 Oakland University\\
Rochester Hills, MI, USA\\
{\tt\small \{shaibalsaha,lxu\}@oakland.edu}
}
\begin{document}
\maketitle
\begin{abstract}
Hybrid models that combine convolutional and transformer blocks offer strong performance in computer vision (CV) tasks but are resource-intensive for edge deployment. \highlight{Although post-training quantization (PTQ) can help reduce resource demand,} its application to hybrid models remains limited. We propose \HyQuant, a novel \search PTQ approach that applies uniform quantization to convolutional blocks and \logtbas quantization to transformer blocks. \HyQuant achieves $2.5\times$–$8.7\times$ latency reduction with minimal accuracy loss on ImageNet-1K dataset. It further demonstrates low latency and memory efficiency on edge devices, making it practical for real-world deployment.
\end{abstract}
\section{Introduction}
The rapid advancements of computer vision (CV) tasks have been driven by convolutional neural networks (CNNs) and, more recently, transformers. While CNNs excel at capturing local features, they struggle with global context~\cite{krizhevsky2017imagenet,naseer2021intriguing,saha2025vision}. Transformers address this through self-attention~\cite{dosovitskiy2020image}, but at the cost of high computational complexity~\cite{8957353, maaz2022edgenext}.

To combine the strengths of both architectures, hybrid models have emerged, integrating convolutional and transformer blocks~\cite{mehta2021mobilevit,mehta2022separable,li2022efficientformer,li2022rethinking}. Figure~\ref{fig:hybrid_1} illustrates architectures with convolutional layers followed by transformer encoders~\cite{li2022bevformer}, whereas Figure~\ref{fig:hybrid_2} uses transformer encoders with convolutional decoders~\cite{hatamizadeh2022unetr}. \highlight{While demonstrating strong performance,} hybrid models remain computationally demanding and \highlight{challenging} to deploy on resource-constrained edge devices~\cite{cai2019once}.

To enable the deployment of hybrid models on edge devices, quantization has emerged as a promising compression technique. While quantization-aware training (QAT) can reduce accuracy loss~\cite{ave2022quantization,gholami2022survey,choi2018pact,jung2019learning}, its high retraining cost makes post-training quantization (PTQ) more practical for real-time inference~\cite{jacob2018quantization,choukroun2019low,zhao2019improving,lee2018quantization}. However, existing PTQ methods designed for CNNs or transformers often struggle to generalize to hybrid architectures~\cite{lee2024q}.

\begin{figure}[]
    \centering
    \begin{minipage}{0.49\linewidth}
        \centering
        \includegraphics[width=\linewidth]{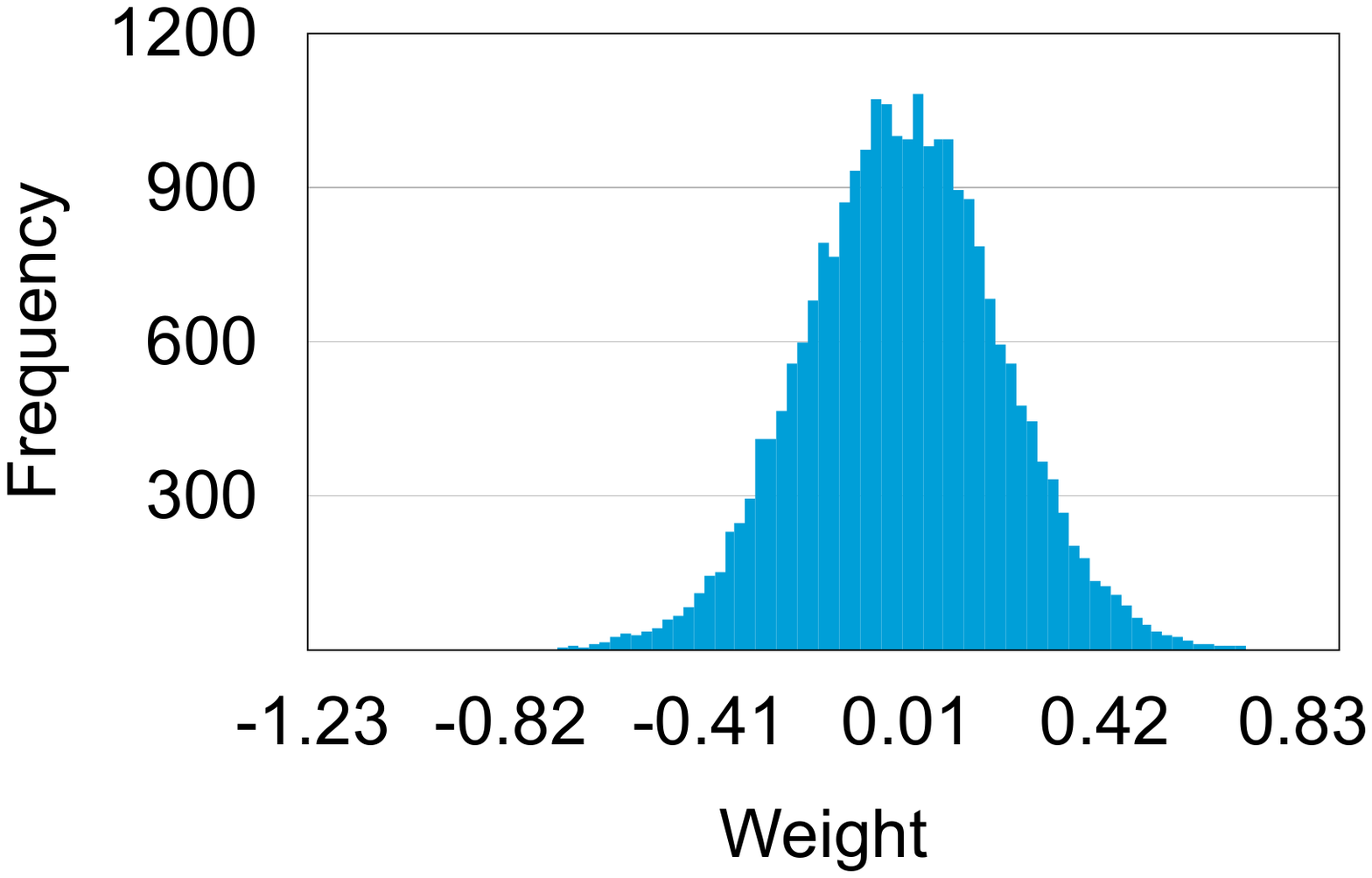}
        \caption{Weight distribution of a convolutional block on MobileViT\_s model.}
        \label{fig:weight_conv}
    \end{minipage}
    \hfill
    \begin{minipage}{0.49\linewidth}
        \centering
        \includegraphics[width=\linewidth]{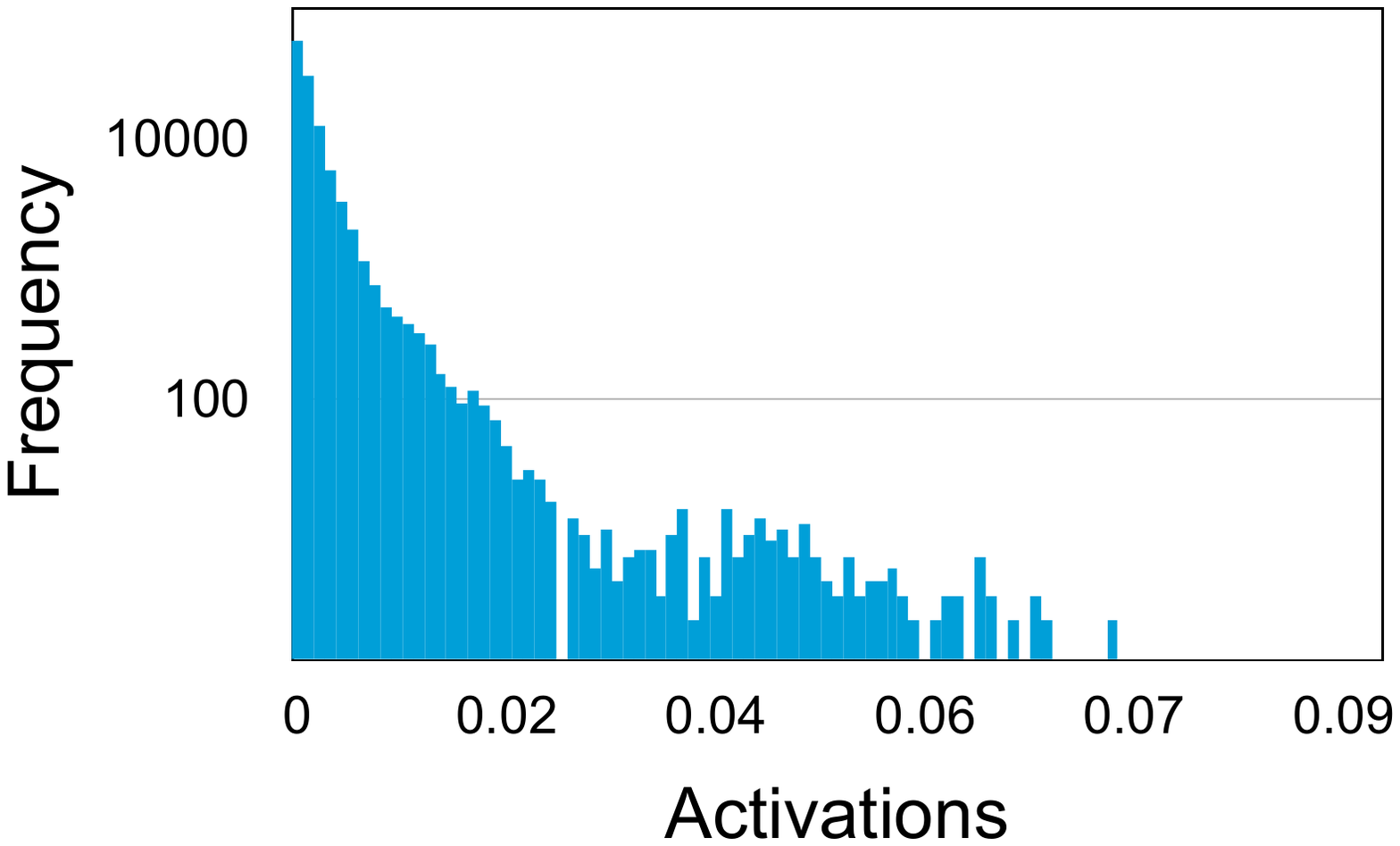}
        \caption{Post-Softmax activation distribution of a transformer block on MobileViT\_s model.}
        \label{fig:activations}
    \end{minipage}
\end{figure}

Recent PTQ methods for hybrid models attempt to bridge the gap but face limitations in generalizability, hardware efficiency, and reliance on large calibration sets. Q-HyViT~\cite{lee2024q} selected granularity and schemes via Hessian-based analysis, but its mixed strategies are not edge-friendly. HyQ~\cite{hyq} proposed quantization-aware distribution scaling (QADS) for convolutional blocks and an integer-only softmax for transformers, \highlight{yet lack evaluation on resource-constrained edge devices such as Jetson Nano}. In contrast, we observe that weights in convolutional block follow a uniform distribution (see Figure~\ref{fig:weight_conv}), while transformer post-Softmax activations follow a power-law pattern (Figure~\ref{fig:activations}), motivating our use of uniform and log$_2$-based quantization, respectively—both efficient for edge deployment~\cite{wu2024adalog,habi2021hptq}. Quantizing both weights and activations in convolutional blocks degrades accuracy, so we apply weight-only quantization. In transformers, weight quantization causes high performance loss, whereas post-Softmax activations tolerate quantization well. Thus, we quantize weights in convolutional blocks and activations in transformer blocks. Based on this, we choose weight quantization for convolutional blocks and activation quantization for transformer blocks. Based on these observations, we propose \HyQuant, a \search PTQ method that automatically identifies and quantizes convolutional and transformer blocks in hybrid models. \HyQuant achieved significant latency improvements and low memory usage on edge devices such as Jetson Nano AGX Xavier, making it practical for real-world deployment. Our main contributions are as follows:
\begin{figure}[]
    \centering
    \begin{subfigure}[b]{0.35\textwidth}
        \centering
        \includegraphics[width=\textwidth]{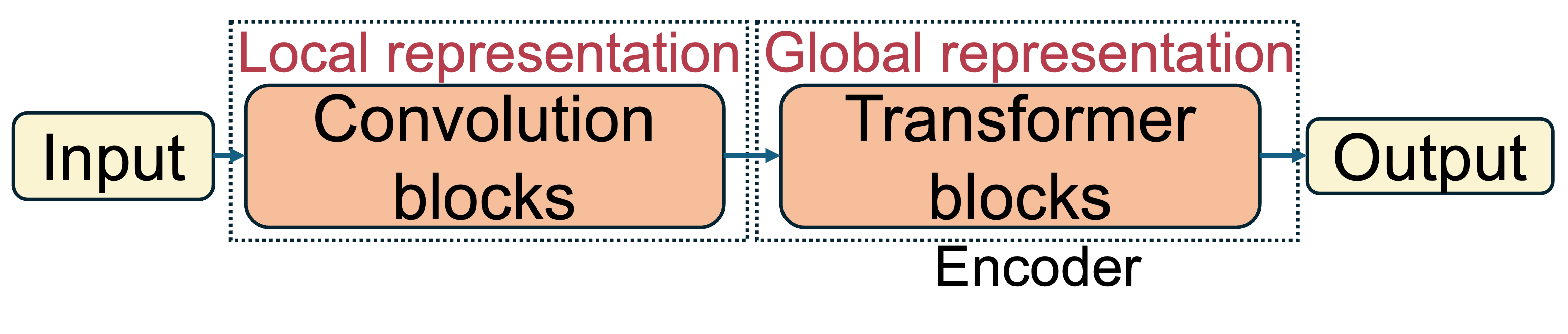}
        \caption{CNN blocks followed by transformer encoder.}
        \label{fig:hybrid_1}
    \end{subfigure}%
    \quad
    \begin{subfigure}[b]{0.35\textwidth}
        \centering
        \includegraphics[width=\textwidth]{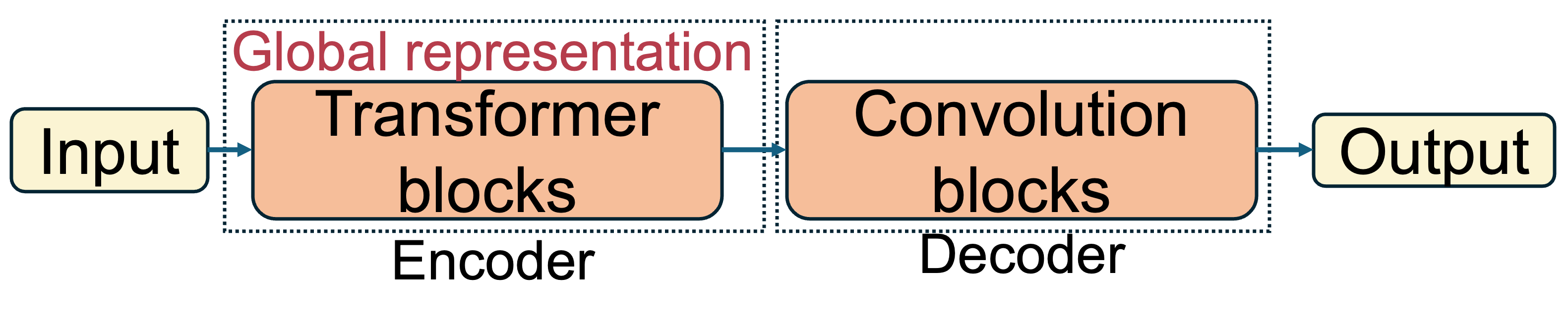}
        \caption{Transformer encoder followed by CNN decoder.}
        \label{fig:hybrid_2}
    \end{subfigure}
    \caption{High-level structures of hybrid models~\cite{mehta2021mobilevit,hatamizadeh2022unetr,li2022bevformer}.}
    \label{fig:hybrid}
\end{figure}
\begin{itemize} 

\item We propose a  \search algorithm to identify convolutional and transformer blocks in hybrid models.

\item We introduce \HyQuant, a PTQ method that applies uniform quantization to convolutional blocks and log$_2$-based quantization to transformer activations, tailored to their distinct characteristics. 

\item \HyQuant achieves lower latency than existing PTQ methods with acceptable accuracy drop and evaluates \HyQuant on edge devices such as Jetson Nano and AGX Xavier. 
\end{itemize}
\section{\HyQuant: PTQ for Hybrid Models}
\begin{figure}[]
    \centering
    \includegraphics[scale=0.3]{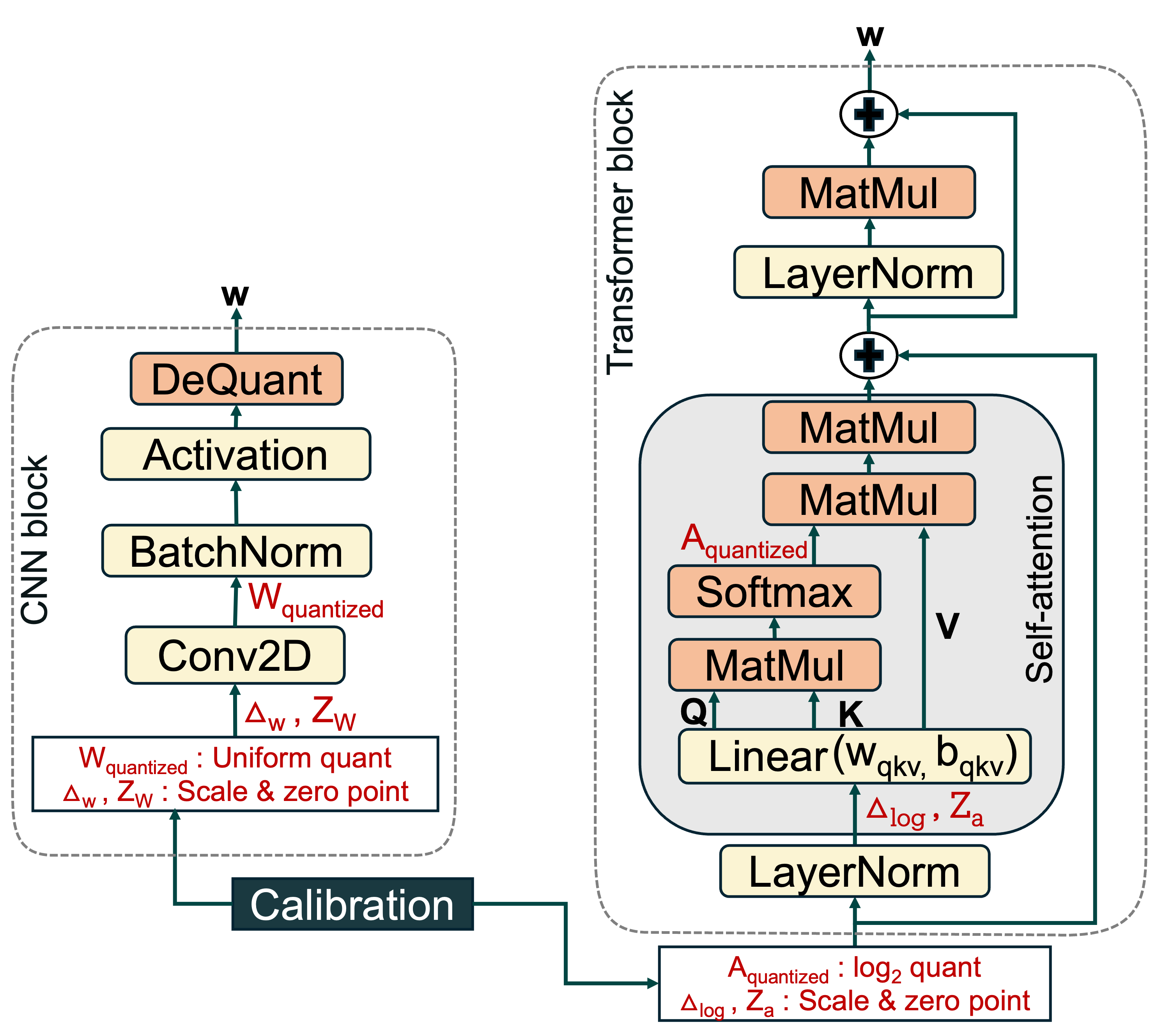}
    \caption{The overview of the \HyQuant to perform uniform and \logtbas quantization for convolution and transformer blocks in the hybrid model.}
    \label{fig:general_overview}
\end{figure}
\noindent \textbf{Overview} Figure~\ref{fig:general_overview} presents an overview of \HyQuant. The following sections detail the \search algorithm, uniform quantization for convolution blocks, and \logtbas quantization for transformer blocks.
\subsection{Structure-Aware Block Identification}\label{search_al_sec} 
We develop a specialized \search algorithm to automatically identify and classify blocks as CNN or transformer within hybrid models, enabling efficient quantization. The algorithm traverses the model using a depth-first search (DFS) with a queue (Q) to systematically explore the architecture (see Algorithm~\ref{search_algorithm}). At each step, it dequeues a module, and its parent identifier, iterates through its sub-blocks, and constructs full identifiers by combining parent and child names. The algorithm returns two lists—$CNN$ and $Transformer$—to guide type-specific quantization. The time complexity of the proposed algorithm is $ \mathcal{O}(n) $. Although implemented in PyTorch, it can be adapted to frameworks like TensorFlow or Keras with minimal adjustments in block-type declarations.
\begin{algorithm}[]
\small
\caption{Structure-Aware Block Identification}\label{search_algorithm}
\textcolor{keywordcolor}{\KwIn{Model $M$}}
\textcolor{keywordcolor}{\KwOut{Lists $CNN$, $Transformer$}}
\textcolor{keywordcolor}{Initialize $Q$, lists $CNN$, $Transformer$\;}\BlankLine
\textcolor{keywordcolor}{Enqueue $(M, "")$ into $Q$\;}\BlankLine
\While{$Q$ is not empty} {
\Comment{Dequeue the last element from the queue}
    $(module, parent\_name) \gets$ dequeue $Q$\;\BlankLine
    \ForEach{$(layer\_name, layer)$ in $module.children()$} {
        $fullname \gets parent\_name + layer\_name$\;\BlankLine
        \If{$layer$ is \texttt{Conv2d}} {
            Append $fullname$ to $CNN$\;
        }
        \ElseIf{$layer$ is \texttt{Linear},"attn","Attention", or "Transformer"} {
            Append $fullname$ to $Transformer$\;
        }
        Enqueue $(layer, fullname)$ into $Q$\;
    }
}
\Return{$CNN$, $Transformer$}\;
\end{algorithm}

\subsection{Uniform Quantization for Convolution Blocks}\label{conv} As we illustrated in Figure~\ref{fig:weight_conv}, the weights in convolution block exhibit uniform distributions. We apply uniform quantization to convolution blocks (stored in the list $CNN$), using fixed scaling factors derived from the block-specific min-max statistics ($[W_{\text{min}}, W_{\text{max}}]$) collected during calibration as like Equation~\ref{zer_s}. 
\begin{equation}
\label{zer_s}
\small
 {\Delta_{\text{W}}} = \frac{W_{\text{max}} - W_{\text{min}}}{2^b - 1},
 \quad Z_{\text{W}} = \text{round}\left( \frac{-W_{\text{min}}}{\Delta_{\text{W}}} \right)
\end{equation}
\begin{equation}
\label{main_eq}
\small
    W_{\text{quantized}} = \text{clamp}\left(\left\lfloor \frac{\text{W} - W_{\text{min}}}{\Delta_W} + Z_{\text{W}} \right\rceil \right)
\end{equation}
\begin{equation}
\label{dequan_un}
\small
W_{\text{dequantized}} = (W_{\text{quantized}} - Z_{W}) \times \Delta_{W} \end{equation}
\highlight{In Equation~\ref{main_eq}}, $W_{\text{quantized}}$ represents 8-bit convolutional weights quantized from FP32 inputs $W \in \mathbb{R}^{\alpha \times \beta \times \gamma}$, where $\alpha$ is the number of channels and $\beta \times \gamma$ are spatial dimensions. The scaling factor $\Delta_{W}$ maps values from $[W_{\text{min}}, W_{\text{max}}]$ to [0, 255], and the zero point $Z_{W}$ \highlight{(From Equation~\ref{zer_s})} aligns the minimum value. Quantized weights are then clamped to the valid \highlight{$b-bit$} range \highlight{(in our method $b=8$)}.

Once the model processes inputs through quantized blocks, weights are restored to floating-point precision during inference for meaningful interpretation. We wrapped the model with conditional dequantization specifically for convolutional blocks, using stored $W_{\text{min}}$ and $W_{\text{max}}$ along with the scaling factor and zero point to calculate the dequantized weights $W_{\text{dequantized}}$ using Equation~\ref{dequan_un}. 
\subsection{\logtbas Quantization for Transformer blocks}\label{transformer_b} 
Based on the understanding from Figure~\ref{fig:activations}, we apply \logtbas quantization to post-Softmax activation in the self-attention mechanisms on transformer blocks (listed in $Transformer$), mapping values onto a logarithmic scale. This preserves small yet critical activations, achieving better precision than traditional methods.

We target 8-bit quantization, using the same clamping range ([0, 255]) as in uniform quantization. Since logarithmic quantization cannot directly handle zero values ($\log(0)=-\infty$), we introduce a small constant $\epsilon=1e^{-5}$ to ensure numerical stability, effectively handling tensors containing zero or very small values.

Since we apply \logtbas quantization on post-Softmax activations (0$ \leq \text{activation} \leq 1$), we only need to handle positive and zero values in transformer blocks. Our quantization approach involves three main components in the transformer blocks: calculating the quantization scaling factor $\Delta_{\text{log}}$, determining the zero point $Z_{\text{a}}$, and applying logtbas-based quantization with clamping. First, the scaling factor $\Delta_{\text{log}}$ determines the intervals for quantizing the values based on the statistics during calibration. Then, we calculate the $Z_{\text{a}}$, which adjusts the quantized values to align with the minimum observed value, ensuring both small and large values are represented within the quantization range. To determine the quantization parameters in the log domain, we first compute:
\vspace{-2mm}
\[
A_{\text{min}} = \log_{2}(a_{\min}), \quad A_{\text{max}} = \log_{2}(a_{\max})
\]
Here, \( a_{\min} \) and \( a_{\max} \) are the minimum and maximum activation values observed during calibration. Based on the $\Delta_{\text{log}}$ and $Z_{\text{a}}$, we apply \logtbas quantization on the post-Softmax activation $a$. Finally, the clamp function ensures that the quantized values $A_{\text{quantized}}$ remain within the valid b-bit range [$0, 2^b - 1$] (see Equation~\ref{dequan_aa}).
\begin{equation}
\small
    \Delta_{\log} = \frac{A_{\text{max}} - A_{\text{min}}}{2^b - 1},
\quad
Z_{\text{a}} = \text{round} \left( \frac{-A_{\text{min}}}{\Delta_{\text{log}}} \right)
\end{equation}
\begin{equation}
  A_{\text{quantized}}
  = \operatorname{clamp}\!\Bigl(\bigl\lfloor\tfrac{-\log_{2}(a + \epsilon)}{\Delta_{\log}} + Z_{a}\bigr\rceil,\;0,\;2^{b}-1\Bigr)
  \label{dequan_aa}
\end{equation}
Where $A_{\text{quantized}} \in \mathbb{R}^{\alpha \times \beta \times \gamma}$ denotes the b-bit integer value; $\alpha$ is the number of channels and $\beta \times \gamma$ is the height and width of the input. The dequantizer $A_{\text{dequantized}}$ can be defined using Equation~\ref{dequan}.
\begin{equation}
\small
\label{dequan}
    A_{\text{dequantized}} =
    2^{(A_{\text{quantized}} - Z_{\text{a}}) \cdot \Delta_{\log}} 
\end{equation}

\section{Experiment}
\subsection{Implementation}\label{implementation_l}
\noindent \textbf{Models and Datasets} We evaluate \HyQuant using MobileViT and MobileViTv2 variants on the ImageNet-1K~\cite{imagenet} validation set. We conduct comprehensive comparisons against state-of-the-art \highlight{(SOTA)} PTQ methods, including EasyQuant~\cite{tang2024easyquant}, PTQ4ViT~\cite{yuan2022ptq4vit}, RepQ-ViT~\cite{li2023repq}, Q-HyViT~\cite{lee2024q}, and HyQ~\cite{hyq}, evaluating our method in terms of accuracy, latency, and memory consumption.

\noindent \textbf{Hardware and Libraries} We deploy MobileViT and MobileViTv2 on RTX 3080, AGX Xavier, and Jetson Nano to evaluate \HyQuant in terms of accuracy, memory consumption, and latency. \HyQuant is implemented in PyTorch and accelerated with TensorRT~\cite{nvidia2021tensorrt} on edge devices. We use pretrained weights from timm~\cite{rw2019timm} library.\\
\noindent \textbf{Experiment Settings} We use batch size 32 for calibration and batch size 1 for edge device inference. For \logtbas quantization, we set $\epsilon = 1e^{-5}$ based on trial and error. Smaller $\epsilon$ increases quantization errors, while larger $\epsilon$ distorts small values which makes \(10^{-5}\) the optimal compromise.

\subsection{Comparison with SOTA PTQ Approaches}
\begin{table*}[]
\centering
\caption{Top-1 (\%) accuracy of hybrid models on ImageNet-1K~\cite{imagenet} after applying PTQ methods originally proposed for pure CNNs, Vision Transformers (ViTs), and hybrid models. ‘–’ indicates unavailable data.}
\label{accuracy_baseline}
\resizebox{0.9\textwidth}{!}{%
\begin{tabular}{c|c|ccc|ccccc}
\hline
\textbf{Models} &
  \textbf{FP32 (\%)} &
  \multicolumn{3}{c|}{\textbf{PTQ for Pure CNN/ ViTs (\%)}} &
  \multicolumn{5}{c}{\textbf{PTQ For Hybrid Models (\%)}} \\ \cline{3-10} 
 &
   &
  \multicolumn{1}{c|}{\textbf{\begin{tabular}[c]{@{}c@{}}EasyQuant\\ ~\cite{tang2024easyquant}\end{tabular}}} &
  \multicolumn{1}{c|}{\textbf{\begin{tabular}[c]{@{}c@{}}PTQ4ViT\\ ~\cite{yuan2022ptq4vit}\end{tabular}}} &
  \textbf{\begin{tabular}[c]{@{}c@{}}RepQ-ViT\\ ~\cite{li2023repq}\end{tabular}} &
  \multicolumn{1}{c|}{\textbf{\begin{tabular}[c]{@{}c@{}}Q-HyViT\\ ~\cite{lee2024q}\end{tabular}}} &
  \multicolumn{1}{c|}{\textbf{\begin{tabular}[c]{@{}c@{}}Q-HyViT\\ (Reproduced)\end{tabular}}} &
  \multicolumn{1}{c|}{\textbf{\begin{tabular}[c]{@{}c@{}}HyQ\\ ~\cite{hyq}\end{tabular}}} &
  \multicolumn{1}{c|}{\textbf{\begin{tabular}[c]{@{}c@{}}HyQ\\ (Reproduced)\end{tabular}}} &
  \textbf{\begin{tabular}[c]{@{}c@{}}\HyQuant\\ (Ours)\end{tabular}} \\ \hline
MobileViT\_xxs &
  68.91 &
  \multicolumn{1}{c|}{36.13} &
  \multicolumn{1}{c|}{37.75} &
  1.85 &
  \multicolumn{1}{c|}{68.20} &
  \multicolumn{1}{c|}{62.73} &
  \multicolumn{1}{c|}{68.15} &
  \multicolumn{1}{c|}{67.94} &
  66.86 \\ \hline
MobileViT\_xs &
  74.62 &
  \multicolumn{1}{c|}{73.16} &
  \multicolumn{1}{c|}{65.52} &
  41.96 &
  \multicolumn{1}{c|}{74.31} &
  \multicolumn{1}{c|}{69.37} &
  \multicolumn{1}{c|}{73.99} &
  \multicolumn{1}{c|}{73.99} &
  73.8 \\ \hline
MobileViT\_s &
  78.31 &
  \multicolumn{1}{c|}{74.21} &
  \multicolumn{1}{c|}{68.19} &
  59.01 &
  \multicolumn{1}{c|}{77.92} &
  \multicolumn{1}{c|}{75.71} &
  \multicolumn{1}{c|}{77.93} &
  \multicolumn{1}{c|}{77.93} &
  77.87 \\ \hline
MobileViTv2\_050 &
  70.1 &
  \multicolumn{1}{c|}{66.80} &
  \multicolumn{1}{c|}{39.39} &
  26.60 &
  \multicolumn{1}{c|}{69.89} &
  \multicolumn{1}{c|}{64.99} &
  \multicolumn{1}{c|}{69.16} &
  \multicolumn{1}{c|}{--} &
  65.59 \\ \hline
MobileViTv2\_075 &
  75.57 &
  \multicolumn{1}{c|}{62.91} &
  \multicolumn{1}{c|}{65.54} &
  55.52 &
  \multicolumn{1}{c|}{75.29} &
  \multicolumn{1}{c|}{69.81} &
  \multicolumn{1}{c|}{74.47} &
  \multicolumn{1}{c|}{--} &
  72.50 \\ \hline
MobileViTv2\_100 &
  77.9 &
  \multicolumn{1}{c|}{69.34} &
  \multicolumn{1}{c|}{51.02} &
  40.61 &
  \multicolumn{1}{c|}{77.63} &
  \multicolumn{1}{c|}{69.66} &
  \multicolumn{1}{c|}{-} &
  \multicolumn{1}{c|}{--} &
  72.03 \\ \hline
MobileViTv2\_125 &
  79.64 &
  \multicolumn{1}{c|}{77.31} &
  \multicolumn{1}{c|}{67.39} &
  41.65 &
  \multicolumn{1}{c|}{79.31} &
  \multicolumn{1}{c|}{71.72} &
  \multicolumn{1}{c|}{-} &
  \multicolumn{1}{c|}{--} &
  77.58 \\ \hline
MobileViTv2\_150 &
  80.36 &
  \multicolumn{1}{c|}{75.83} &
  \multicolumn{1}{c|}{68.61} &
  62.12 &
  \multicolumn{1}{c|}{79.97} &
  \multicolumn{1}{c|}{73.09} &
  \multicolumn{1}{c|}{-} &
  \multicolumn{1}{c|}{--} &
  79.08 \\ \hline
MobileViTv2\_175 &
  80.86 &
  \multicolumn{1}{c|}{79.93} &
  \multicolumn{1}{c|}{72.30} &
  63.52 &
  \multicolumn{1}{c|}{80.45} &
  \multicolumn{1}{c|}{75.20} &
  \multicolumn{1}{c|}{-} &
  \multicolumn{1}{c|}{--} &
  80.30 \\ \hline
MobileViTv2\_200 &
  81.12 &
  \multicolumn{1}{c|}{80.04} &
  \multicolumn{1}{c|}{75.50} &
  64.65 &
  \multicolumn{1}{c|}{80.76} &
  \multicolumn{1}{c|}{75.85} &
  \multicolumn{1}{c|}{-} &
  \multicolumn{1}{c|}{-} &
  80.48 \\ \hline
\end{tabular}%
}
\end{table*}
\noindent \textbf{Accuracy} Table~\ref{accuracy_baseline} compares the top-1 accuracy of \HyQuant with PTQ methods designed for CNNs (EasyQuant), ViTs (PTQ4ViT, RepQ-ViT), and hybrid models (Q-HyViT, HyQ) on the ImageNet-1K dataset. The results for EasyQuant, PTQ4ViT, and RepQ-ViT are adopted from Q-HyViT and show significant accuracy degradation when applied to hybrid models due to architecture-specific design. Our implementation of Q-HyViT on RTX 3080 yields lower accuracy than originally reported, while HyQ closely matches its original results. Although HyQ achieves the highest top-1 accuracy overall, it does not support complex MobileViTv2 variants (MobileViTv2\_100 to MobileViTv2\_200). In contrast, \HyQuant maintains stable accuracy across all models, with slightly higher degradation in smaller variants (e.g., 6.86\% on MobileViTv2\_050) and minimal degradation in larger variants (e.g., 0.69\% on MobileViTv2\_175). In some scenarios like MobileViT\_s, \HyQuant performs comparably to Q-HyViT. These results highlight \HyQuant robustness and scalability across diverse hybrid architectures.

\begin{figure}[]
    \centering
    \includegraphics[width=\linewidth]{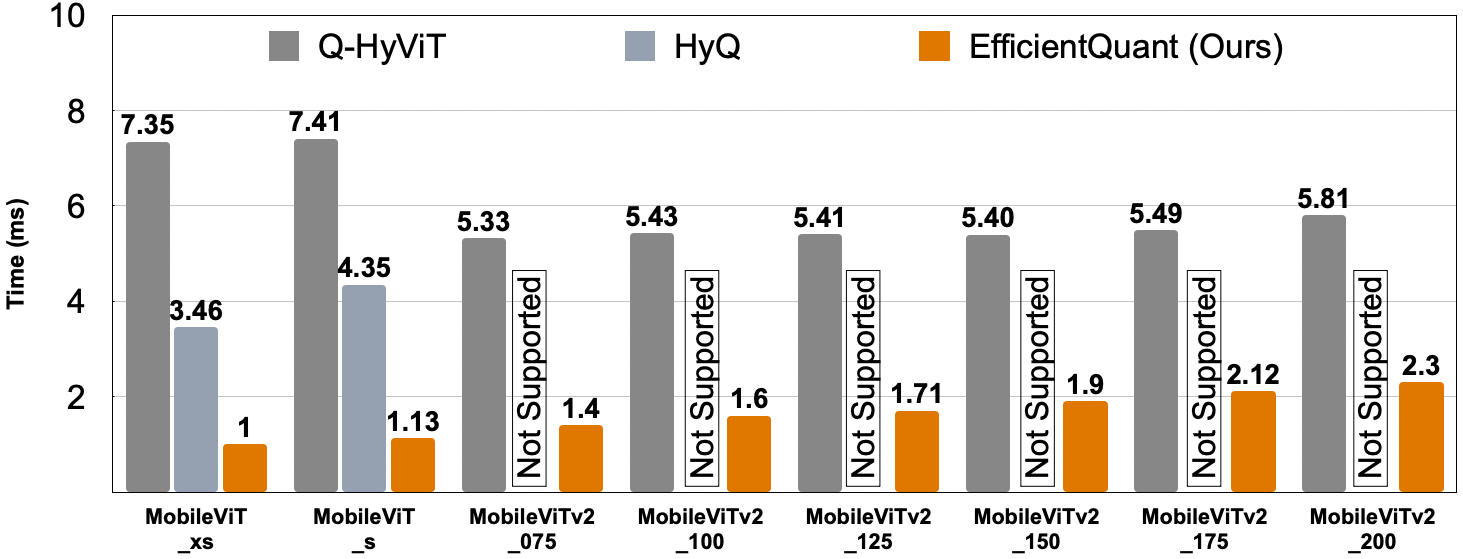}
    \caption{\HyQuant achieves better latency on RTX 3080 GPU than the SOTA techniques on MobileViT and MobileViTv2 models. }
    \label{fig:accuracy_sota}
\end{figure}

\noindent \textbf{Latency} \HyQuant outperforms Q-HyViT and HyQ in inference latency on RTX 3080 across all MobileViT variants. For fairness, we re-evaluated all methods on the RTX 3080 GPU. As shown in Figure~\ref{fig:accuracy_sota}, HyQ supports only smaller models, while \HyQuant scales to the full MobileViTv2 range. Notably, \HyQuant achieves a \textbf{$2.5\times$} - \textbf{$8.7\times$} latency reduction over Q-HyViT, with MobileViTv2\_175 running at  \textbf{2.36 ms} vs \textbf{5.81 ms}. Compared to HyQ, \HyQuant delivers up to \textbf{$3.85\times$} faster inference (e.g., \textbf{1.13 ms} vs. \textbf{4.35 ms} on MobileViT\_s) while maintaining comparable accuracy. These results highlight \HyQuant’s efficiency, scalability, and practicality for edge deployments.
\subsection{Evaluation with Edge Devices}
\begin{figure}[]

    \centering
    \includegraphics[width=\linewidth]{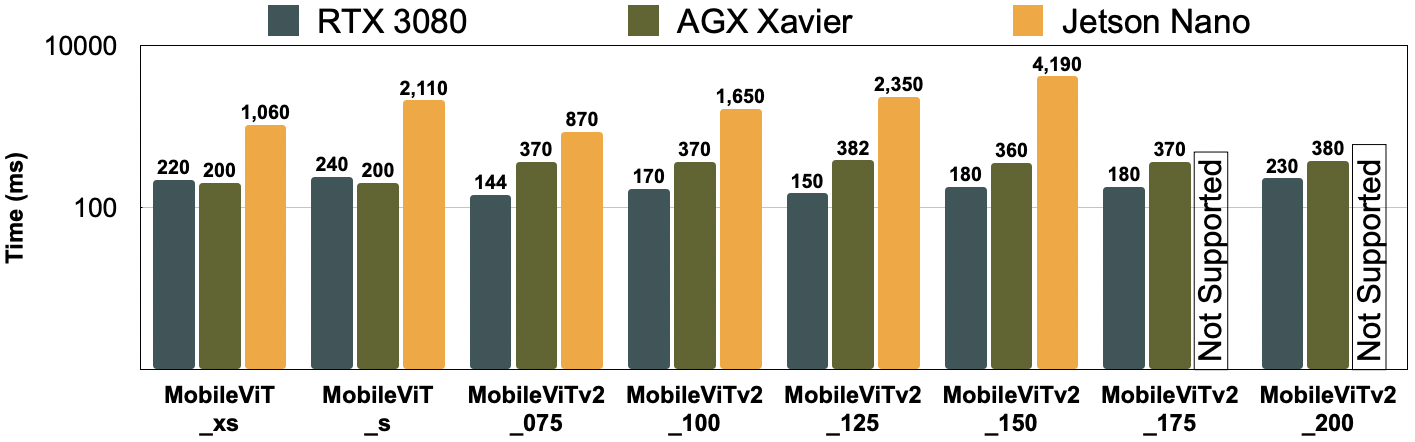}
    \caption{Load time of MobileViT models on edge devices with \HyQuant using TensorRT engine.}
    \label{fig:load_time}
\end{figure}

\begin{figure}[]

    \centering
    \includegraphics[width=\linewidth]{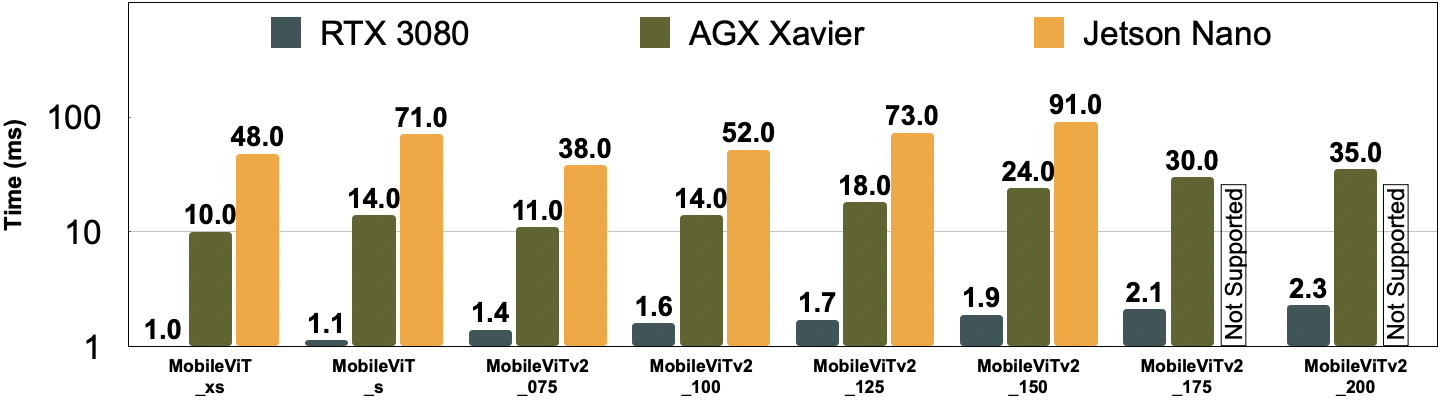}
    \caption{Inference time of MobileViT models on edge devices with \HyQuant using TensorRT engine.}
    \label{fig:edge_pure}
\end{figure}

\noindent \textbf{Latency} We evaluate \HyQuant on edge devices based on TensorRT engine load and inference time. Since no prior work has explored hybrid model deployment on the edge device, baseline comparisons are not available. 
Figure~\ref{fig:load_time} illustrates the engine load time for MobileViT variants. The average model load times for the RTX 3080, AGX Xavier, and Jetson Nano are 187.4 ms, 320.2 ms, and 1477.0 ms, respectively, which are acceptable for edge deployment. Figure~\ref{fig:edge_pure} presents inference time comparisons for \HyQuant across RTX 3080, AGX Xavier, and Jetson Nano. Jetson Nano exhibits significantly higher latency, achieving $3\times$ to $5\times$ slower than AGX Xavier and up to $63\times$ slower than RTX 3080. The average inference times—1.51 ms on RTX 3080, 40.7 ms on AGX Xavier, and 17.8 ms on Jetson Nano—demonstrate acceptable latency for edge deployment. Based on our observations, Jetson Nano is better suited for static deployments given its higher load times. In contrast, AGX Xavier offers a balanced trade-off between efficiency and flexibility, making it well-suited for CV tasks.

\noindent \textbf{Memory} We also measure the maximum memory usage during the inference times for all three devices. We utilize \textit{pynvml} for RTX 3080 and \textit{tegrastats} library for AGX Xavier and Jetson Nano for measuring memory usage. Each validation set was run ten times to obtain an average maximum memory usage per model. Results indicated slight variations among the devices, with average memory usage being 2,568 MiB for the RTX 3080, 2,344 MiB for AGX Xavier, and 2,265 MiB for the Jetson Nano. The overall differences in memory usage are $\sim2\%$ because of different memory management across those devices.
\section{Conclusion}
In this study, we proposed \HyQuant, a novel \search PTQ method for hybrid models that automatically identifies block types to enable targeted quantization. Leveraging the observed distribution patterns in convolutional and transformer blocks, \HyQuant applies uniform and \logtbas quantization accordingly. It achieves up to $8.7\times$ latency reduction with minimal accuracy loss. Evaluations on RTX 3080, AGX Xavier, and Jetson Nano validate the efficiency and adaptability of \HyQuant, highlighting its suitability for low-latency deployment on a wide range of resource-constrained edge devices.

\section*{Acknowledgements}
This work was supported in part by the \highlight{U.S.} National Science Foundation under Grant No. 2245729, and by the University Research Committee Faculty Research Fellowship at Oakland University.
\small
    \bibliographystyle{ieeenat_fullname}
    \bibliography{reference}


\end{document}